\documentclass[runningheads]{llncs}

\usepackage{eccv}

\usepackage{eccvabbrv}

\usepackage{graphicx}
\usepackage{booktabs}

\usepackage{amsthm}
\usepackage{booktabs}
\usepackage{amssymb}
\usepackage{amsmath} 
\usepackage{multirow}
\usepackage[countmax]{subfloat}

\usepackage{enumitem}

\usepackage[accsupp]{axessibility}  

\usepackage[pagebackref,breaklinks,colorlinks,citecolor=eccvblue]{hyperref}

\begin{document}

\title{Towards RGB-NIR Cross-modality Image Registration and Beyond} 


\author{
Huadong Li$^{1,*}$ \and
Shichao Dong$^{1,*}$ \and
Jin Wang$^{1,*}$ \and
Rong Fu$^{1,*}$ \and \\
Minhao Jing$^{1}$ \and
Jiajun Liang$^{1}$ \and
Haoqiang Fan$^{1}$ \and
Renhe Ji$^{1,\dag}$
}

\authorrunning{H. Li et al.}

\institute{$^{1}$MEGVII Technology \\
\email{ \{lihuadong, furong, jirenhe\}@megvii.com \\
\{dongshichao1996, wjbillbieber\}@gmail.com} 
}

\def\footnotesymbollist{%
  \symbol{$*$}%
  \symbol{$\dag$}%
}

{
	\renewcommand{\thefootnote}{\fnsymbol{footnote}}
     \footnotetext[1]{Equal contribution}
	\footnotetext[4]{Corresponding author}
}

\maketitle

\begin{abstract}
   This paper focuses on the area of RGB(visible)-NIR(near-infrared) cross-modality image registration, which is crucial for many downstream vision tasks to fully leverage the complementary information present in visible and infrared images.
   In this field, researchers face two primary challenges - the absence of a correctly-annotated benchmark with viewpoint variations for evaluating RGB-NIR cross-modality registration methods and the problem of inconsistent local features caused by the appearance discrepancy between RGB-NIR cross-modality images. 
   To address these challenges, we first present the RGB-NIR Image Registration (RGB-NIR-IRegis) benchmark, which, for the first time, enables fair and comprehensive evaluations for the task of RGB-NIR cross-modality image registration.
   Evaluations of previous methods highlight the significant challenges posed by our RGB-NIR-IRegis benchmark, especially on RGB-NIR image pairs with viewpoint variations.
   To analyze the causes of the unsatisfying performance, 
   we then design several metrics to reveal the toxic impact of inconsistent local features between visible and infrared images on the model performance.
   This further motivates us to develop a baseline method named Semantic Guidance Transformer (SGFormer), which utilizes high-level semantic guidance to mitigate the negative impact of local inconsistent features.
   Despite the simplicity of our motivation, extensive experimental results show the effectiveness of our method.
   \emph{The code and the benchmark will be released when the paper is accepted.}  
   \keywords{RGB-NIR cross-modality \and Image registration \and Benchmark}

\end{abstract}

\section{Introduction}


\begin{figure*}[t]
\centering
\includegraphics[width=0.9\textwidth]{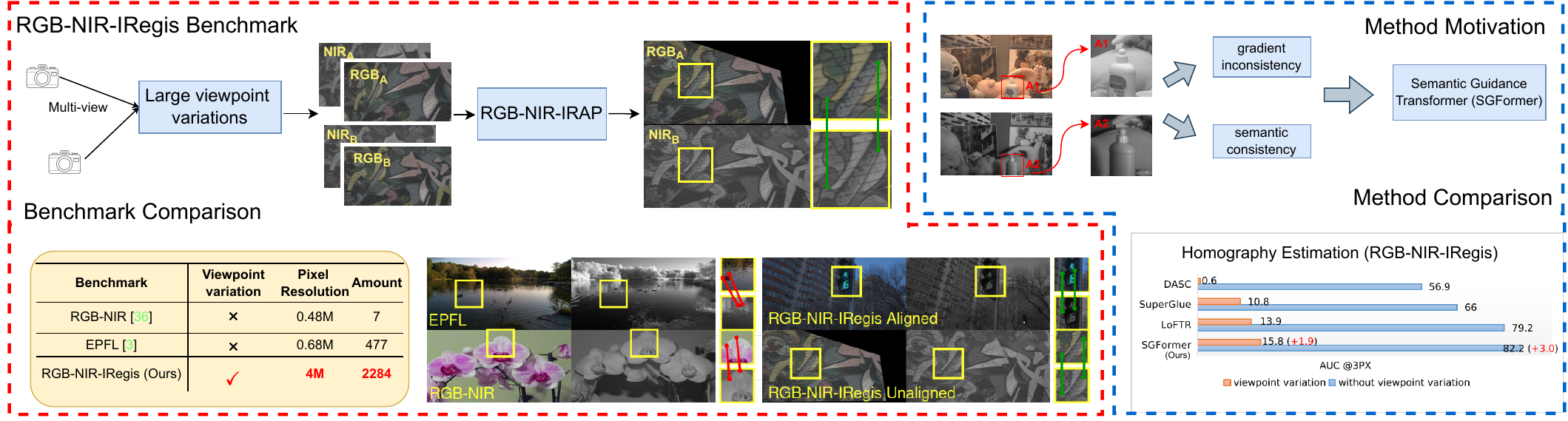}
\caption{\textbf{Overview of our framework.}  
As shown in the left part of the figure, with the proposed \textbf{RGB-NIR} \textbf{I}mage \textbf{R}egistration \textbf{A}nnotation \textbf{P}ipeline (dubbed \textbf{RGB-NIR-IRAP}), we present the largest-scale \textbf{RGB-NIR} \textbf{I}mage \textbf{Regis}tration (dubbed \textbf{RGB-NIR-IRegis}) benchmark featuring \emph{correct annotations} (green lines) and \emph{viewpoint variations} (unaligned RGB-NIR image pairs).
This is in comparison to previous benchmarks \cite{shen2014multi,brown2011multi}, where factually correct matching correspondences are annotated as errors (red lines) and RGB-NIR image pairs are limited to restricted viewpoint variations.
Utilizing the proposed RGB-NIR-IRegis benchmark, we fairly show that existing registration methods perform poorly, \emph{e.g}, DASC \cite{kim2015dasc,kim2016dasc} obtained AUC @3px of only 0.6 when assessed in the RGB-NIR cross-modality and cross-viewpoint scenario.
This motivates us to analyze the causes for such poor performance and further develop a baseline method called SGFormer, which achieved new SOTA results for the task of RGB-NIR cross-modality image registration.
}
\label{fig:benchmark}
\end{figure*}



In recent years, RGB(visible)-NIR(near-infrared) cross-modality has received an increasing research focus in a broad range of vision tasks, \emph{e.g.}, 24-hour surveillance \cite{surveillance}, autonomous driving \cite{autonomousdriving}, object detection \cite{objectdetection}, person re-identification (ReID) \cite{reID1,reID2} and low-light imaging \cite{Scalemap,jin2022darkvisionnet}.
In these tasks, leveraging RGB-NIR cross-modality images gathers a wider range of environmental information, which helps to overcome the inherent limitations of a single modality \cite{Zhang2018ADA,Zhang2019FeatherNetsCN,Zhi2019MultispectralIF,Salamati2012SemanticIS,Martin2019DriveActAM}.

However, in practical scenarios, RGB and NIR images are often sourced from distinct devices or sensors, resulting in inherent misalignment between image pairs, such as parallaxes. 
This easily overlooked misalignment issue can pose \textit{undesired yet significant} challenges to state-of-the-art downstream vision algorithms, impeding their abilities to attain superior performance.
For instance, neglecting the misalignment of RGB-NIR image pairs can severely damage the performance of state-of-the-art image fusion methods \cite{xu2022rfnet}.
Therefore, to achieve optimal performance in downstream tasks, \textit{RGB-NIR cross-modality image registration} becomes crucial, which aims to obtain dense cross-modality correspondences between RGB and NIR images.

However, it is worth noting that the field of RGB-NIR cross-modality image registration remained largely underexplored in the past. 
Specifically, in contrast to the well-explored field of general image registration \cite{rublee2011orb,detone2018superpoint,sarlin2020superglue,sun2021loftr,edstedt2022dkm}, which focuses on estimating correspondences between two visible images, we identify two primary challenges in this domain that significantly impede its progress.
\begin{enumerate}[leftmargin=*]
\item[$\bullet$] \emph{Challenge 1.} To the best of our knowledge, there still lacks a \textbf{correctly-annotated} RGB-NIR cross-modality image registration benchmark to enable \textbf{fair} evaluations for current registration methods.
\end{enumerate}

Collecting a comprehensive image registration benchmark with correct annotations for RGB-NIR image pairs has posed a long-standing challenge. 
Firstly, manually annotating such dense visual correspondences between cross-modality image pairs requires considerable human effort. 
Besides, there also lacks a robust RGB-NIR cross-modality image registration method to support pre-annotations in advance to alleviate human efforts.
Consequently, previous studies \cite{shen2014multi,brown2011multi} have been limited to annotating only a small number of RGB-NIR image pairs, often with restricted viewpoint variations and susceptible to human error.
Evaluating on these benchmarks casts doubt on the design effectiveness of registration methods, which further hinders improvements for future studies.

To mitigate these issues, we propose the \textbf{RGB}-\textbf{NIR} \textbf{I}mage \textbf{R}egistration \textbf{A}nnotation \textbf{P}ipeline (dubbed \textbf{RGB-NIR-IRAP}) to effectively provide correct annotations for a large number of aligned and unaligned RGB-NIR image pairs (\emph{i.e.}, RGB-NIR image pairs without and with viewpoint variations).
Specifically, RGB-NIR-IRAP utilizes a monocular RGB-NIR sensor \cite{tang2015high} to first acquire naturally pixel-wise aligned RGB-NIR image pairs.
To further annotate correspondences for RGB-NIR image pairs with viewpoint variations, our proposed RGB-NIR-IRAP then constructs RGB-NIR image quadruplets and innovatively converts the cross-modality image registration annotation task into the single-modality image registration annotation task (RGB-RGB, NIR-NIR). 
Subsequently, by leveraging the experience of single-modality image registration annotation procedures from HPatches \cite{balntas2017hpatches}, our proposed RGB-NIR-IRAP can effectively annotate correct and dense correspondences for a large number of RGB-NIR image pairs.

With the proposed RGB-NIR-IRAP, we then present the \textbf{RGB}-\textbf{NIR} \textbf{I}mage \textbf{Regis}tration (dubbed \textbf{RGB-NIR-IRegis}) benchmark, which, for the first time,  supports fair and reliable evaluations of current registration methods on RGB-NIR image pairs. 
To the best of our knowledge, \textit{this is the largest-scale RGB-NIR image registration benchmark to date, encompassing RGB-NIR image pairs with correct annotations and significant viewpoint variations}.
Qualitative and quantitative comparisons with previous benchmarks \cite{brown2011multi,shen2014multi} are shown in Fig. \ref{fig:benchmark}.
Utilizing our newly-proposed RGB-NIR-IRegis benchmark, we then perform comprehensive evaluations of previous methods.
As shown in Fig. \ref{fig:benchmark}, the performance of these methods is far from satisfactory, particularly when assessed on RGB-NIR image pairs with viewpoint variations. 
These findings highlight the challenges of the RGB-NIR-IRegis benchmark and can effectively guide future research directions for further enhancements. 
We believe that our RGB-NIR-IRegis benchmark will serve as a pivotal platform to nurture innovative studies in this critical domain.

Moreover, this evaluation results further motivate us to explore the reasons for the poor performance of previous registration methods, which reveals another unique challenge in this domain.
\begin{enumerate}[leftmargin=*]
\item[$\bullet$] \emph{Challenge 2.} The \textbf{appearance discrepancy} between visible and infrared images causes \textbf{inconsistent local features} of RGB-NIR image pairs, which requires critical attention for registration methods. 
\end{enumerate}
To demonstrate this, we first design several metrics to quantify the impact of inconsistent local features on the performance of previous methods.
Specifically, we focus on the \emph{RGB-NIR gradient inconsistency} to measure such inconsistencies between RGB and NIR images (See Fig. \ref{fig2:benchmark-sample} for examples.).
Experimental results show that such inconsistencies can lead to previous methods (\emph{e.g.}, SIFT \cite{lowe1999object}) matching the wrong corresponding locations.
To address this challenge, we further develop a baseline method named Semantic Guidance Transformer (SGFormer) based on our analyses.
Our intuition is that humans typically rely on the high-level semantic representations of images (\emph{e.g.}, visual concepts) to find correspondences in RGB-NIR cross-modality image pairs, where the same object may appear differently.
Despite the simplicity of the motivation, we show that by introducing both global and local semantic guidance, 
SGFormer successfully outperformed SOTA methods in RGB-NIR cross-modality image registration, effectively alleviating the local feature inconsistency issue.

\section{Related Work}
In this paper, we roughly divide previous studies into general image registration and cross-modality image registration.

\noindent \textbf{General Image Registration.} 
General image registration methods aimed to generate correspondences between two visible images.
Roughly speaking, there were two main categories for general image registration methods: detector-based methods and detector-free methods.
Traditional detector-based methods 
SIFT \cite{lowe1999object} and ORB \cite{rublee2011orb} were proposed to match local features across different images depending on detected key points.
Deep learning based self-supervised methods SuperPoint \cite{detone2018superpoint} and Mishkin \textit{et al.} \cite{mishkin2018repeatability} learned reliable key points and descriptors at the pixel level without annotated benchmarks.
Semi-supervised methods 
SuperGlue \cite{sarlin2020superglue} and Wang \textit{et al.} \cite{wang2020learning} established correspondences using camera pose relations relying on the large-scale SfM reconstruction benchmarks.
Detector-free methods LoFTR \cite{sun2021loftr} and COTR \cite{jiang2021cotr} proposed to improve registration performance by directly generating correspondences in raw images.
On this basis, ASpanFormer \cite{chen2022aspanformer} and Chen \emph{et al.} \cite{chen2022guide} further performed refinement on the results of coarse matching.
However, given the circumstance of RGB-NIR image pairs with evident appearance discrepancy, such methods often suffered from significant performance drops.

\noindent \textbf{Cross-modality Image Registration.}  Previously analyzed by Irani and Anandan \cite{irani1998robust}, the differences between multi-spectral images have posed significant challenges to cross-modality image registration.
To begin with, EPFL \cite{brown2011multi} and RGB-NIR \cite{shen2014multi} benchmarks were proposed to support the evaluation for cross-modal image registration methods despite the slight misalignment issue.
To this end, LGHD \cite{lghd2015}, ANCC \cite{heo2010robust} and DASC \cite{kim2015dasc,kim2016dasc} were proposed to find cross-modality robust features.
However, the above methods were usually time-consuming and sparse-matching.
Arar \emph{et al.} \cite{arar2020unsupervised} employed an unsupervised learning image-to-image translation to convert the NIR modality to the RGB modality, aiming to bypass the cross-modality issue.
Deep learning based unsupervised stereo matching methods \cite{liang2019unsupervised,chen2022degradation,walters2021there,liang2022deep} attempted to synthesize cross-modality images for pixel-level dense matching.
Moreover, Tosi \emph{et al.} \cite{tosi2022rgb} and Zhi \emph{et al.} \cite{zhi2018deep} proposed RGB-NIR stereo matching dataset for training and evaluation.
Besides, Zhou \emph{et al.} \cite{zhou2022promoting} designed a self-supervised framework to promote the single-modal optical flow networks for diverse cross-modality registration.
However, the above cross-modality methods and benchmarks mainly focused on the cases of limited geometric transformations, lacking the evaluation for various geometric differences across modalities.
In contrast, our proposed RGB-NIR-IRegis benchmark contains images of diverse viewpoint variations and real scenes with multi-illumination, which provides a comprehensive and challenging benchmark for future studies.

\section{RGB-NIR Image Registration Benchmark}\label{benchmark}
In this section, we first present the \textbf{RGB}-\textbf{NIR} \textbf{I}mage \textbf{Regis}tration (dubbed \textbf{RGB-NIR-IRegis}) benchmark, which aims to support fair and comprehensive evaluations of registration methods on RGB-NIR cross-modality image pairs.
Specifically, our proposed RGB-NIR-IRegis benchmark encompasses both aligned and unaligned RGB-NIR image pairs with correct dense visual correspondence annotations, which is effectively constructed with 
our proposed \textbf{RGB}-\textbf{NIR} \textbf{I}mage \textbf{R}egistration \textbf{A}nnotation \textbf{P}ipeline (dubbed \textbf{RGB-NIR-IRAP}).

To support the fairness of strict evaluations, our RGB-NIR-IRAP first aims to provide pixel-wise correct annotations for the aligned RGB and NIR image pairs (\emph{i.e.}, RGB and NIR image pairs without viewpoint variations).
To this end, different from previous benchmarks \cite{shen2014multi,brown2011multi}, our RGB-NIR-IRegis benchmark collected images with the monocular RGB-NIR sensor \cite{tang2015high}, which captures the scene under the same spatial and temporal condition, naturally ensuring the pixel-wise alignment of the paired RGB and NIR images, thus having accurate annotations.
In comparison, the EPFL \cite{brown2011multi} collected cross-modality image pairs depending on the alternation of RGB and NIR filters.
During the collection of images, small movements of the camera tripod may cause the misalignment between the paired RGB and NIR images.
Besides, due to the sequential imaging time of the same scene, the position of the same object (\emph{e.g.} clouds, animals and etc.) on RGB and NIR images may also vary.
As a result, such unintentional offsets inevitably lead to the evaluation results of image registration methods being both unreliable and unfair.
RGB-NIR \cite{shen2014multi} collected cross-modality image pairs depending on the binocular RGB-NIR sensor with inevitable parallel misalignment.
Although they manually annotated several key points for evaluation, the number of annotated key points was far limited.
In contrast, we collected images with the monocular RGB-NIR sensor.
The camera simultaneously recorded RGB and NIR images with a single sensor, thus being able to produce pixel-wise RGB-NIR aligned cross-modality images at the same time.


\begin{figure}[t]
    \centering
\includegraphics[width=0.8\textwidth]{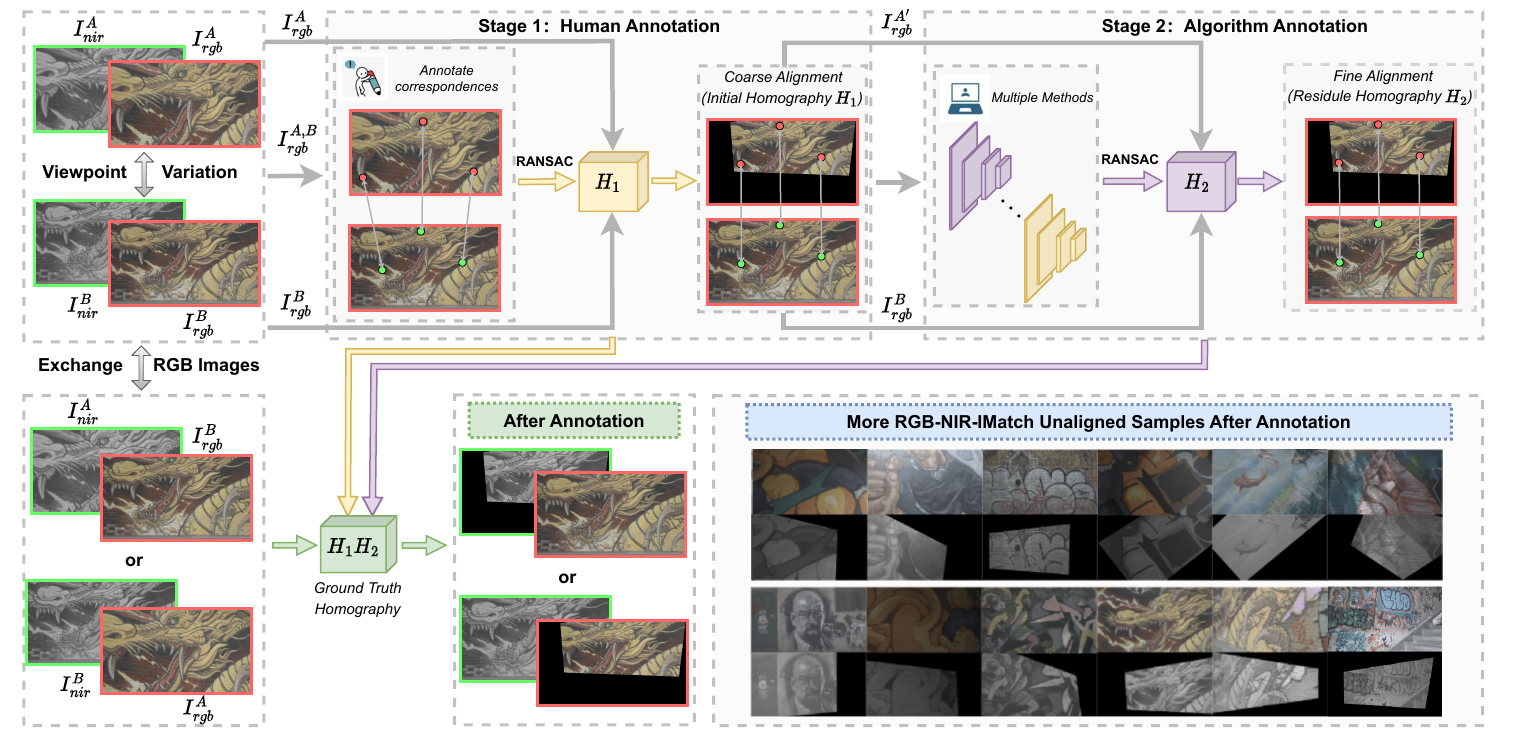}
\caption{The proposed \textbf{RGB}-\textbf{NIR} \textbf{I}mage \textbf{R}egistration \textbf{A}nnotation \textbf{P}ipeline (dubbed \textbf{RGB-NIR-IRAP}) for collecting RGB-NIR image pairs with viewpoint variations.
 }
\label{fig:label_process}
\end{figure}

Moreover, as the unique and significant challenges for image registration, image pairs sharing viewpoint variations play an important role in evaluating previous image registration methods (\emph{cf.} Tab. \ref{tab:RGB-NIR Matching}).
Therefore, our proposed RGB-NIR-IRAP further collected RGB-NIR images with various geometric differences in our benchmark. 
Leveraging the pixel-wise alignment property of our RGB-NIR image pairs, we could obtain the correct correspondences of unaligned RGB-NIR image pairs in an efficient manner.
Specifically, following procedures illustrated in Fig. \ref{fig:label_process}, we first used our monocular RGB-NIR sensor to construct RGB-NIR image quadruplets, \emph{i.e.}, two pairs
of RGB-NIR images with different viewpoints, \emph{i.e.}, pair $A=\{I^A_{rgb}, I^A_{nir}\}$ and $B=\{I^B_{rgb}, I^B_{nir}\}$. 
Then, given the visible images $I^A_{rgb}$ and $I^B_{rgb}$, we followed HPatches \cite{balntas2017hpatches} to use human annotations and algorithm annotations \cite{lowe1999object,sarlin2020superglue,sun2021loftr}, obtaining the initial homograph $H_1$ and the residual homograph $H_2$, respectively.
The ground truth homograph between $I^A_{rgb}$ and $I^B_{rgb}$ was then set as $H_1 H_2$. 
At last, thanks to the pixel-wise alignment property of our aligned RGB-NIR image pairs, the ground truth homograph between unaligned RGB-NIR cross-modality image pairs $I^A_{rgb}$ and $I^B_{nir}$ (or $I^A_{nir}$ and $I^B_{rgb}$) was also $H_1 H_2$. 
We collected 25 scene sequences, containing 260 image pairs under different viewpoints, enriching our benchmark to support more
challenging cross-modality image registration evaluations. 
In addition, to comprehensively evaluate the generalization of RGB-NIR image registration methods, images in our benchmark are ensured to cover multiple scenes with varying conditions.
To this end, we collected images with multi-illumination and higher resolution compared with previous benchmarks. 

For further discussion, directly applying the above procedure to unaligned RGB-NIR image pairs (\emph{e.g.}, $I^A_{rgb}$ and $I^B_{nir}$) might prove impractical.
To begin with, due to the imaging differences between RGB and NIR images, accurately annotating the corresponding pixels between unaligned RGB-NIR image pairs can be difficult and labor-intensive for humans.
Besides, there still lacks a robust RGB-NIR cross-modality image registration methods to facilitate efficient algorithm annotations.
In contrast, by adhering to the procedure in Fig. \ref{fig:label_process}, we could effectively establish the cross-modality correspondences between RGB-NIR unaligned image pairs.
This was achieved by the inherent pixel-wise alignment property of our aligned RGB-NIR image pairs.



Detailed comparisons with previous benchmarks \cite{brown2011multi,shen2014multi} are summarised in Fig. \ref{fig:benchmark}. 
To the best of our knowledge, our proposed RGB-NIR-IRegis benchmark is \textit{the largest-scale RGB-NIR cross-modality image registration benchmark to date, featuring pixel-wise correct annotations and viewpoint variations}.
With our RGB-NIR-IRegis benchmark, we conduct comprehensive evaluations of previous registration methods, showing that their performance falls short on the task of RGB-NIR cross-modality image registration, especially on RGB-NIR image pairs with viewpoint variations (\emph{cf.} Tab. \ref{tab:RGB-NIR Matching}).
This highlights the significant challenges of our proposed RGB-NIR-IRegis benchmark, which we believe can serve as an instrumental platform to foster and monitor the development of this field.

\begin{figure}[t]
\centering
\includegraphics[width=0.8\textwidth]{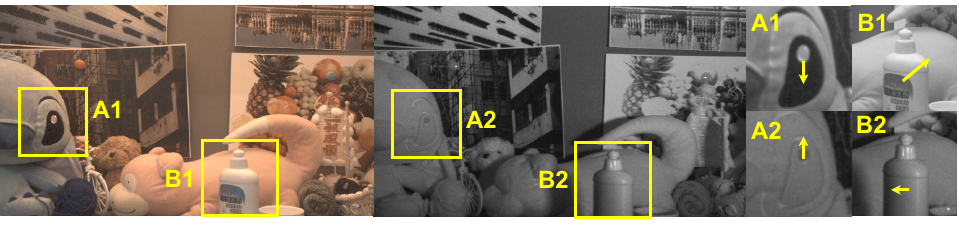}
\caption{\textbf{Examples of \emph{RGB-NIR gradient inconsistency}}. RGB image patches are converted to grayscale for comparison. The yellow arrow represents the gradient of the pixel with its length indicating the magnitude. 
Patches show the inconsistency between RGB-NIR image pairs in the gradient A) orientation; B) both magnitude and orientation.
}
\label{fig2:benchmark-sample}
\end{figure}
\section{RGB-NIR Gradient Inconsistency} \label{gi}
In this section, leveraging the newly proposed RGB-NIR-IRegis benchmark, we aim to further analyze the causes for the poor performance of previous registration methods.
Specifically, we propose to analyze the impact of inconsistent local features on the performance of RGB-NIR cross-modality image registration\footnote{Without the loss of generality, we used aligned image pairs in RGB-NIR-IRegis benchmark in this section for experiments.}, which stems from the appearance discrepancy between visible and infrared images.
For example, the NIR spectrum has stronger penetrating power than visible light, which may make the edges of objects in foggy scenes clearer.
Such differences may lead to the model performance drop during cross-modality image registration.

Specifically, we focus on the inconsistent local features between visible and infrared images from the perspective of \emph{RGB-NIR gradient inconsistency} (See Fig. \ref{fig2:benchmark-sample}.). 
To comprehensively analyze its influence, we first conduct the RGB-NIR varied gradient distribution experiment to verify the existence of gradient inconsistency between the matching patches in the paired RGB-NIR images.
Then, we quantify the impact of gradient inconsistency on current general and cross-modality image registration methods to verify its toxic impact on RGB-NIR cross-modality image registration.

\subsection{RGB-NIR Varied Gradient Distribution}

\begin{figure}[t]
    \centering
    \subfloat[HPatches]{
    \includegraphics[width=0.25\textwidth]{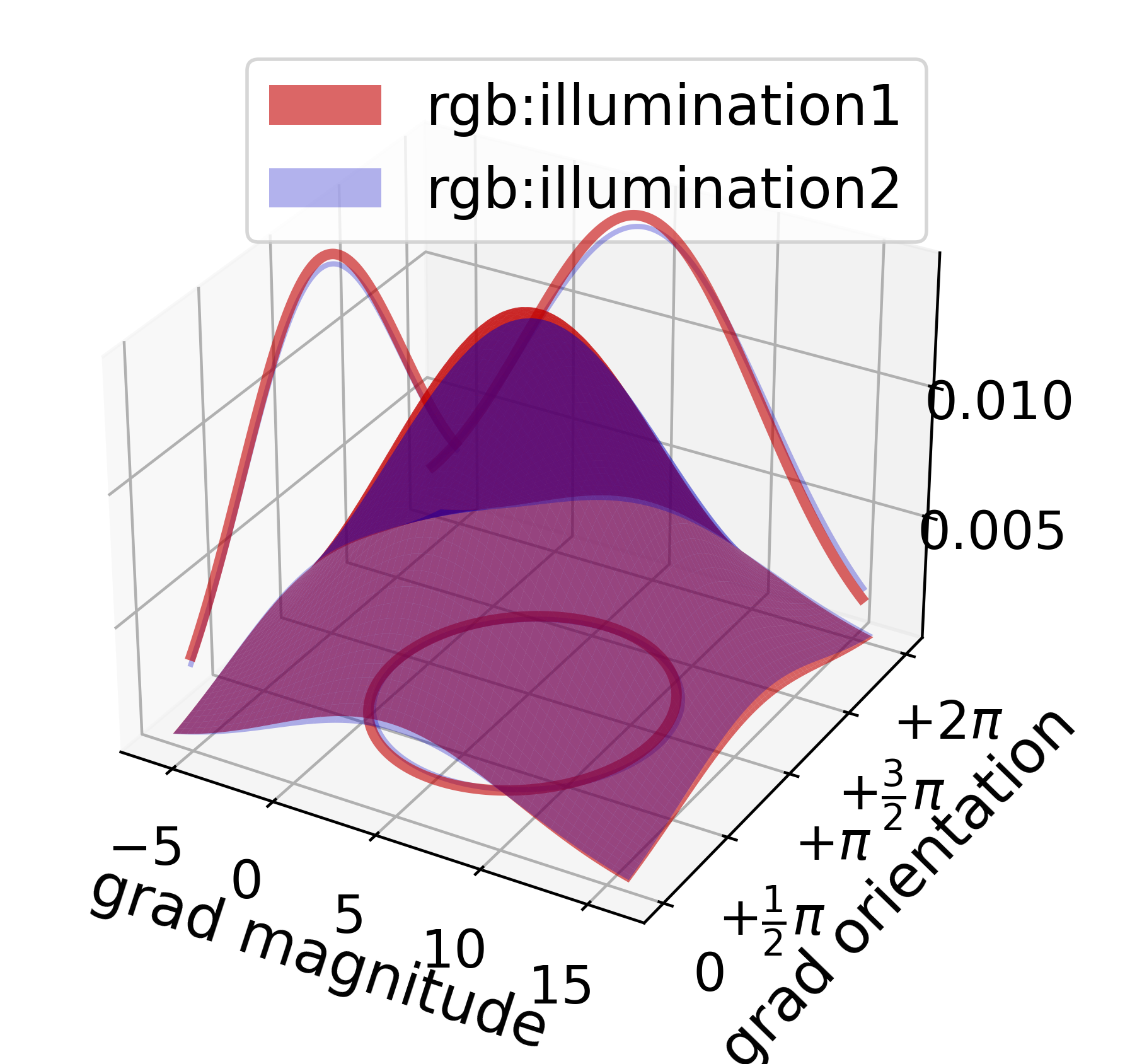}
    \label{Hpatch}}
    \subfloat[EPFL]{
    \includegraphics[width=0.25\textwidth]{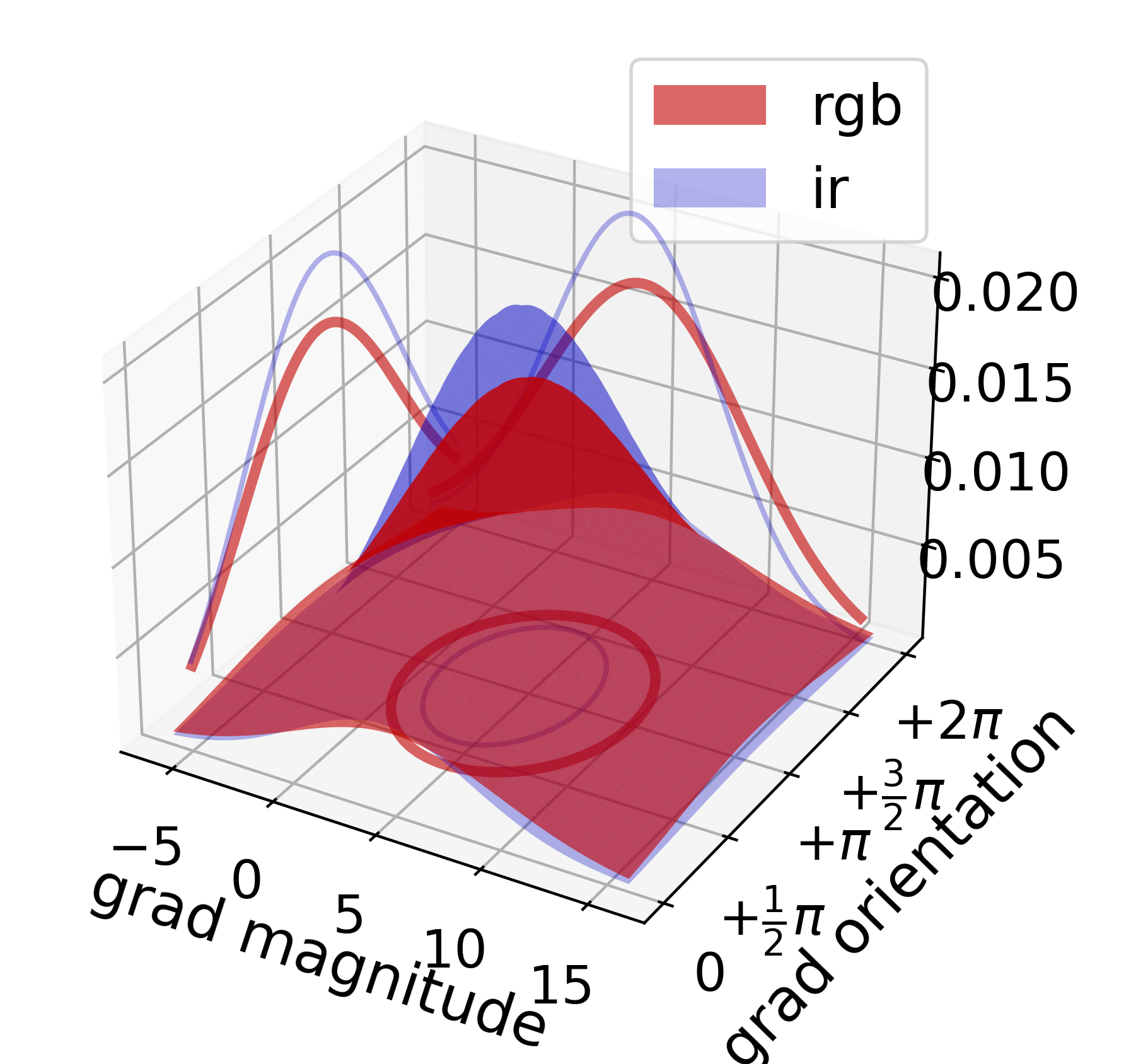}
    \label{EPFL}
     }
    \subfloat[Ours]{
     \includegraphics[width=0.25\textwidth]{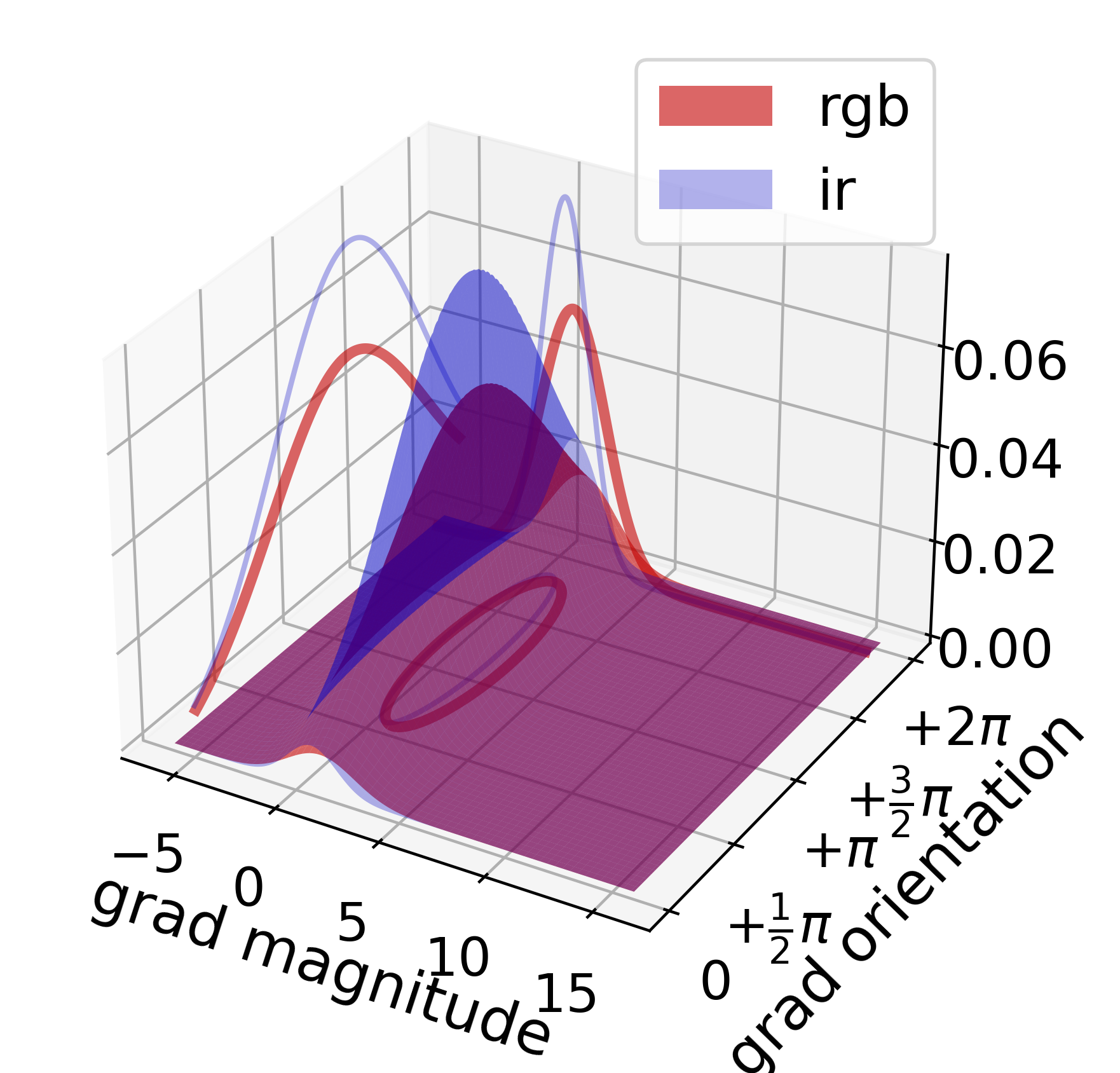}
      \label{RGB-NIR-IMatch}
     }
\caption{\textbf{Gradient distribution comparison using Bivariate Gaussian Distribution maps}. 
Compared with the gradient distribution shift between visible images with illumination variations in the same sensor (Fig. \ref{Hpatch}), there existed significant gradient distribution differences between visible and infrared images (Fig. \ref{EPFL} and \ref{RGB-NIR-IMatch}), which were more evident in our RGB-NIR-IRegis benchmark.
}
\label{fig1:gradient_distribution}
\end{figure}

In this section, we measured the gradient distribution of the overall RGB and NIR images by quantifying the gradient information of local patches.
Similar to SIFT \cite{lowe1999object}, we used the gradient magnitude and gradient orientation to represent the gradient information of a pixel as follows, \emph{i.e.,}
$g_o(i, j) = \arctan(\frac{\partial f(i,j) / \partial i}{\partial f(i,j) / \partial j})$ and
$g_m(i, j) = \sqrt{\frac{\partial f(i,j)}{\partial i}^{2} + \frac{\partial f(i,j)}{\partial j}^{2}}$.
$g_m(i, j)$ and $g_o(i, j)$ indicate the gradient magnitude and gradient orientation of the pixel $(i, j)$ respectively. 
$f(i,j)$ represents the intensity of pixel $(i, j)$.

We further performed patch-based gradient modeling to resist the interference of image noise.
The gradient orientation in a patch was equally divided into $K$ bins.
Then the orientation with the highest frequency among the $K$ subsets was considered as the gradient orientation of the patch.
The sum of the gradient magnitudes in this orientation was considered as the gradient magnitude of the patch.
Specifically, for a given $L \times L$ patch, we got the gradient magnitude $G_{m}$ and gradient orientation $G_{o}$ of a patch as follows.
\begin{equation} \label{eq3}
\small
\begin{split}
\begin{aligned}
G_{m} = \max_k(\sum_{(i,j)\in S_k} g_m(i, j)), 
G_{o} = \mathop{\arg\max_k}(\sum_{(i,j) \in S_{k}}g_m(i, j)) \times \frac{(2 \pi)}{K} 
\end{aligned}
\end{split}
\end{equation}
where $S_{k}$ represents the $k$-th subset of pixels based on the orientation division; $k \in \{ 1,...,K \}$. 
In experiments, we set $L=16, K=8$.

To further illustrate the significant variations between RGB and NIR images, we compared the differences in the gradient distribution between both visible images with illumination variations and cross-modality images.
To this end, we used three benchmarks, including RGB-RGB HPatches \cite{balntas2017hpatches} benchmark (\emph{i.e.},  visible images with illumination variations.), RGB-NIR  EPFL \cite{brown2011multi} benchmark, and our proposed RGB-NIR-IRegis benchmark.
We calculated the mean and variance of gradient metrics $G_m$, $G_o$ in those benchmarks based on patches, and fit them into a Bivariate Gaussian Distribution to roughly compare the distributional differences.
As shown in Figure \ref{fig1:gradient_distribution}, in cross-modality benchmarks, RGB images and NIR images had varied gradient distributions, which were not similar to the differences caused by illumination variations in a single sensor (Fig. \ref{Hpatch}).
Moreover, by collecting a large scale of multi-illumination, multi-scene and pixel-wise correctly-annotated images, our proposed  RGB-NIR-IRegis benchmark further demonstrated more significant gradient inconsistency between visible and infrared images, compared with the previous cross-modality benchmark EPFL \cite{brown2011multi}.

\subsection{RGB-NIR Gradient Inconsistency Impact}

\begin{figure}[t]
    \centering
     \subfloat[\fontsize{8.0pt}{\baselineskip}\selectfont {RGB-NIR cross-modality}]{
     \includegraphics[width=0.35\textwidth]{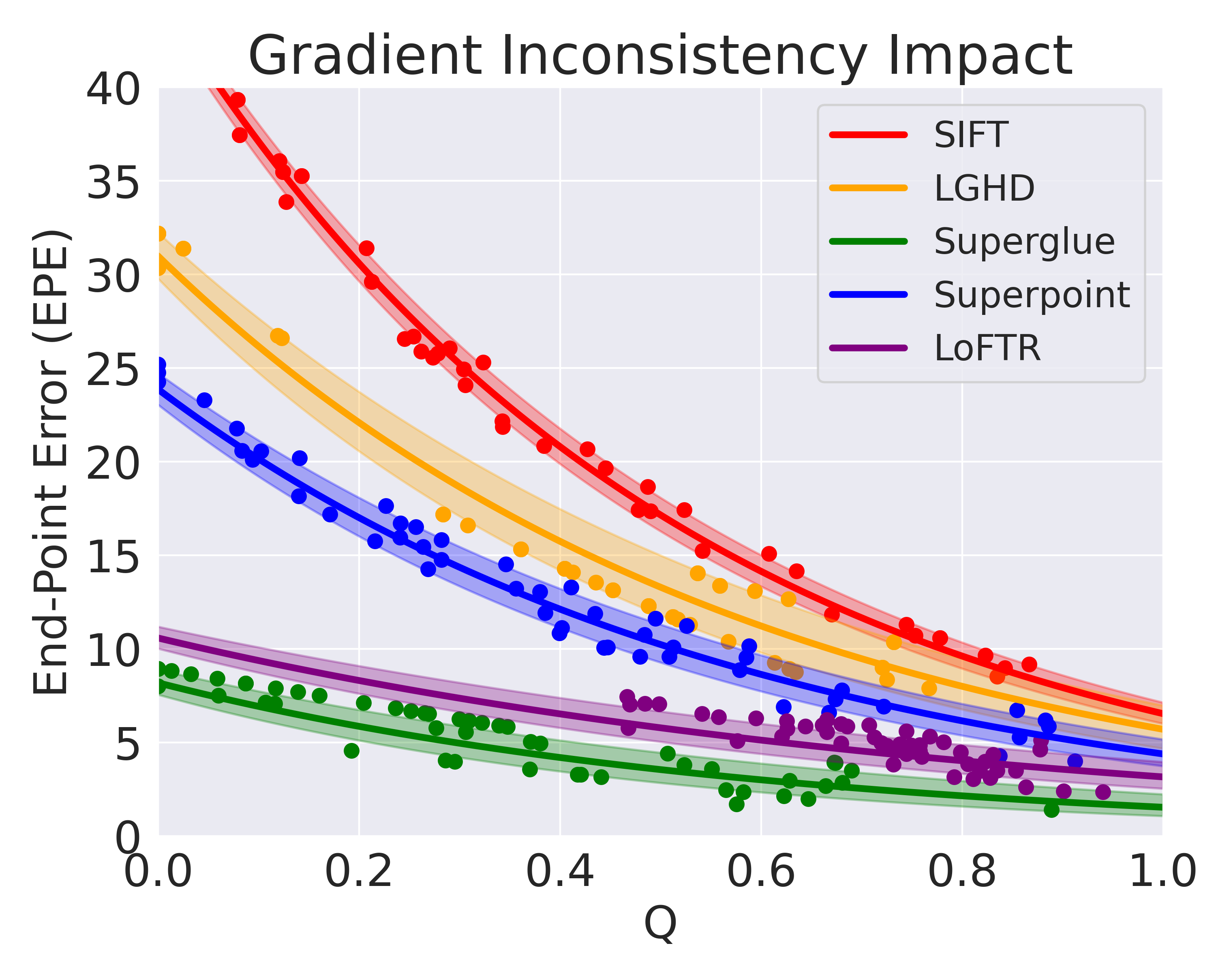}
     }
     \hfill
     \subfloat[\fontsize{8.0pt}{\baselineskip}\selectfont RGB-RGB cross-illumination]{
     \includegraphics[width=0.35\textwidth]{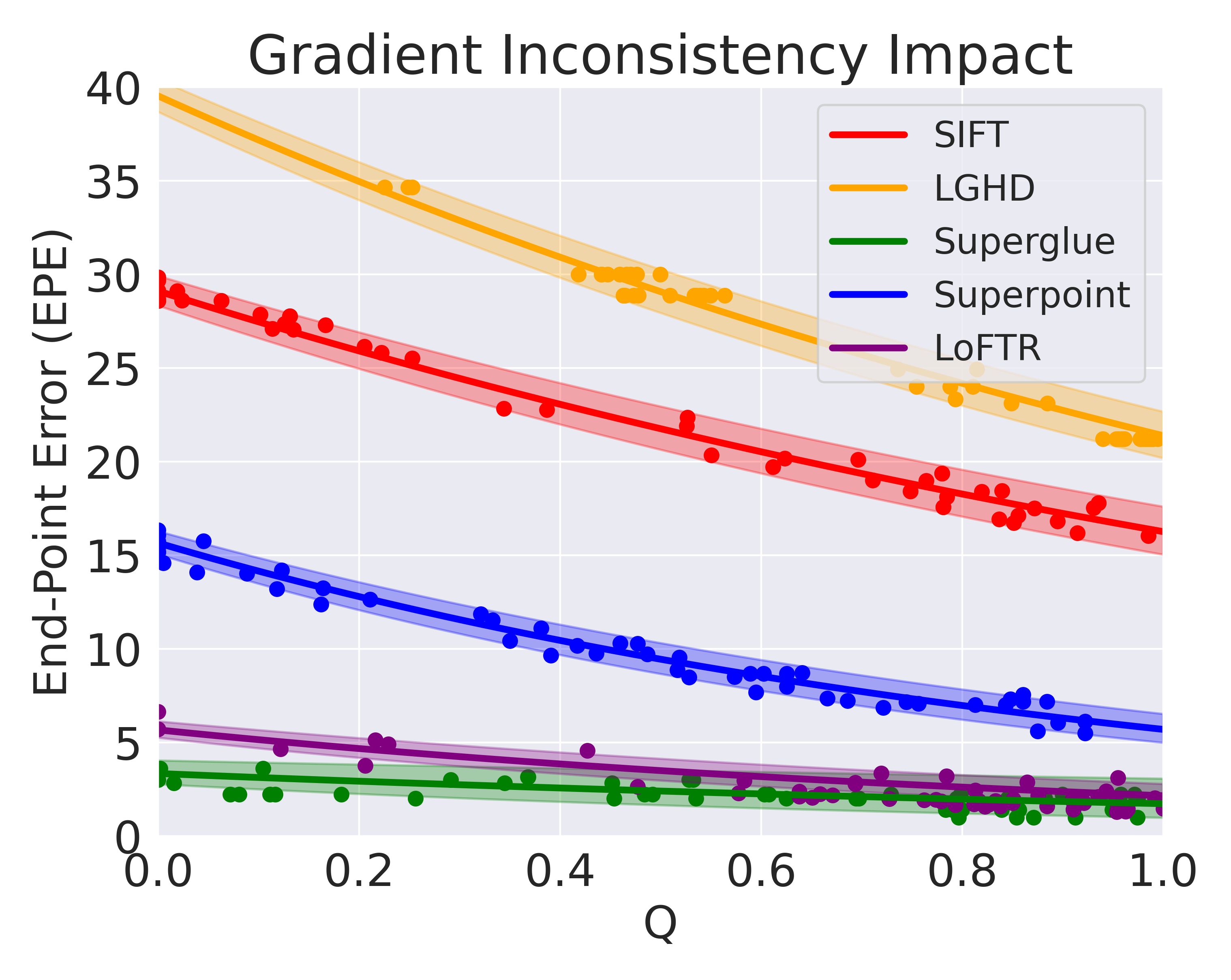}
     }
\caption{\textbf{RGB-NIR gradient inconsistency impact comparisons.} The X-axis and Y-axis represent the gradient inconsistency metric $Q$ and the prediction deviation $EPE$. 
Results show that as the gradient inconsistency increased, the registration performance of current methods declined. 
Besides, the negative impact of gradient inconsistency of RGB-NIR cross-modality (RGB-NIR-IRegis) caused a more significant performance drop trend than that of RGB-RGB cross-illumination (HPatches) in general  (\emph{i.e.}, having sharper slopes).
}
\label{fig2}
\end{figure}

After verifying the existence of RGB-NIR gradient inconsistency in the cross-modality benchmark, we further expect to analyze its impact on the accuracy of various current general and cross-modality image registration methods.

Taking the influence of both gradient magnitude and gradient orientation into account, we first defined a gradient inconsistency quantization metric $Q(p_x, p_y)$ to measure the gradient differences between any matching patches $p_x$ and $p_y$.
$Q(p_x, p_y)$ was calculated as follows.
\begin{equation}\small
\label{eq5}
\begin{split}
\begin{aligned}
Q(p_x, p_y) & = \frac{\min(G_{m}(p_x), G_{m}(p_y))}{\max(G_{m}(p_x), G_{m}(p_y))} 
 \times \frac{\cos(|G_{o}(p_x) - G_{o}(p_y)|) + 1}{2} 
\end{aligned}
\end{split}
\end{equation}
where $G_{m/o}(\cdot)$ represents the gradient magnitude/orientation for the given matching patches $p_x$ and $p_y$ as defined in Eq. \ref{eq3}; $Q \in [0, 1]$.
A larger value of $Q$ indicates more consistent gradient information between the given matching patches.

To measure the performance of current methods, we leveraged the widely-used metric End-Point Error ($EPE$) \cite{kim2015dasc,shen2014multi}, which refers to the average L2-distance between the estimated point coordinates and the ground truth. 

Based on the above metrics, we then evaluated the relationship between the RGB-NIR gradient inconsistency and the registration performance of various methods.
As shown in Figure \ref{fig2}, there existed a negative relationship between the gradient inconsistency metric $Q$ and the prediction deviation $EPE$.
As the gradient inconsistency increased, the performance of the model prediction decreased.
Such results illustrate that current image registration methods depended on gradient features to varying degrees for generating image correspondences.
Moreover, compared to the illumination variations, RGB-NIR cross-modality also presented more challenges to current image registration methods in general (\emph{i.e.}, having sharper slopes).

\section{Semantic Guidance Transformer}
Based on the analysis of the RGB-NIR gradient inconsistency problem, we further propose a baseline method named Semantic Guidance Transformer (SGFormer) to boost the performance of RGB-NIR image registration.

\textbf{Motivation:} 
Analyses in Sec. \ref{gi} show that inconsistent local features between visible and infrared image pairs are insufficient enough to discover correct matching correspondences, which implies that the task of RGB-NIR cross-modality image registration may essentially call for high-level semantic representations as guidance.
Specifically, though visible and infrared images may demonstrate inconsistent gradient information on local patches, they still share common high-level semantic consistency.
To this end, our SGFormer is designed to inject semantic features into descriptors through the Semantic Injection Module (SIM).
By forcing the model to learn semantic features in local patches, the negative impact of the RGB-NIR gradient inconsistency can be reduced.
Moreover, to guide our model to learn the differences between various semantic areas in images, we introduce Semantic Triplet Loss to further boost the registration performance. 

\textbf{Semantic Injection Module.}
Adaptive Instance Normalization (AdaIN) \cite{huang2017arbitrary} has been widely used to integrate high-level semantic information into image representations, such as ID features of human face images \cite{chen2020simswap,faceshifter,facecontroller}.
Inspired by such methods, we design the Semantic Injection Module (SIM) to refine descriptors with semantic features. 

\begin{figure}[t]
\centering
\includegraphics[width=0.8\textwidth]{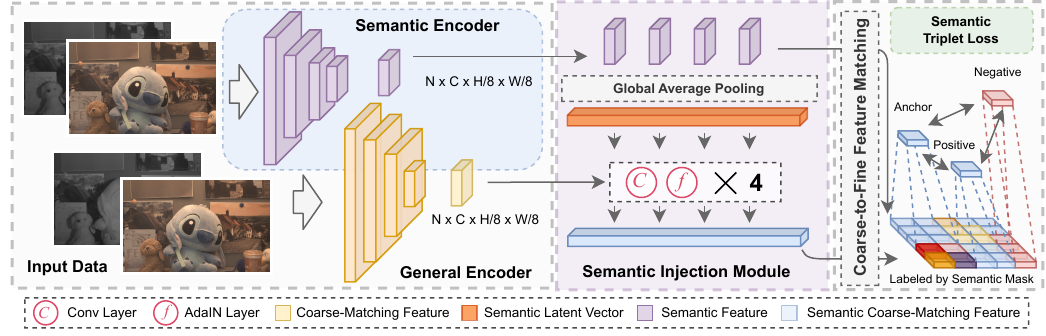}
\caption{\textbf{The overall framework of Semantic Guidance Transformer (SGFormer).} With the guidance of Semantic Injection Module and Semantic Triplet Loss, SGFormer expects to integrate semantic information into the learning of descriptors.}
\label{fig:sgformer}
\end{figure}

The overall framework of the Semantic Injection Module (SIM) is shown in Figure \ref{fig:sgformer}.
The SIM takes descriptors extracted from the general encoder as the input and injects semantic features into descriptors.
Specifically, we exploit a pre-trained semantic encoder \cite{zhou2017scene,zhou2018semantic} to extract semantic features from input images.
Then, the global average pooling is added after the semantic encoder to get the semantic latent vector.
Finally, inspired by the design of the SPADE \cite{SPADE} and AdaIN \cite{huang2017arbitrary}, we refine the descriptors based on the semantic vector as follows:
\begin{equation}\label{EQ:SIM}\small
SIM(\boldsymbol{h}_g,\boldsymbol{h}_s)=\gamma(\boldsymbol{h}_s)\frac{\boldsymbol{h}_g-\boldsymbol{\mu}_{\boldsymbol{h}_g}}{\boldsymbol{\sigma}_{\boldsymbol{h}_g}}+\beta(\boldsymbol{h}_s)  
\end{equation}
where $\boldsymbol{h}_g$ represents the descriptors extracted from the general encoder. 
$\boldsymbol{\mu}_{\boldsymbol{h}_g}$ and $\boldsymbol{\sigma}_{\boldsymbol{h}_g}$ denote the channel-wise mean and standard
deviation of $\boldsymbol{h}_g$.
$\gamma(\boldsymbol{h}_s)$ and $\beta(\boldsymbol{h}_s)$ denote the semantic information extracted from a $1 \times 1$ Conv layer which takes the semantic latent vector $\boldsymbol{h}_s$ as the input.

Overall, the SIM generates refined descriptors with semantic features.
After fusing high-level features, our model learns to generate correspondences adaptively based on the semantic information across RGB and NIR images.
Thus, the interference of the RGB-NIR gradient inconsistency can be reduced.

\textbf{Semantic Triplet Loss.}
To further facilitate the guidance of SIM to refine descriptors, we design the Semantic Triplet Loss.
During the training phase, we get the semantic label of each patch in the input image through the pre-trained semantic model \cite{zhou2017scene,zhou2018semantic}.
Given a semantically refined descriptor $\bar{\boldsymbol{h}}_{a}$, the Semantic Triplet Loss ensures that, the distance between $\bar{\boldsymbol{h}}_{a}$ and a positive descriptor $\bar{\boldsymbol{h}}_{p}$ belonging to the same semantic class should be closer than, the distance between $\bar{\boldsymbol{h}}_{a}$ and a negative descriptor $\bar{\boldsymbol{h}}_{n}$ belonging to another class, by at least a margin $m$.
Moreover, to reduce the impact of the wrong semantic classification, we only select patches of top $T$ classification confidences to calculate the loss.
Inspired by \cite{hermans2017defense}, our Semantic Triplet Loss is shown as follows.
\begin{equation}\small
\begin{split}
\mathcal{L}_{st}(\bar{\boldsymbol{h}}) = \overset{all {\kern 2pt} anchors}{\overbrace{\sum_{c=1}^{C}\sum_{a=1}^{T}  } }  [m + 
\overset{hardest {\kern 1pt} positive}{\overbrace{\underset{p=1, ..., T}{max} D(\bar{\boldsymbol{h}}^{c}_{a}, \bar{\boldsymbol{h}}^{c}_{p})}}  - 
\overset{hardest {\kern 1pt} negative}{\overbrace{\underset{c^{\prime}=1, ..., C,  c^{\prime} \ne c}{\underset{n=1, ..., T}{min} }D(\bar{\boldsymbol{h}}^{c}_{a}, \bar{\boldsymbol{h}}^{c^{\prime}}_{n})}}]_{+} 
\end{split}
\end{equation}
where $D(\cdot, \cdot)$ denotes the distance metric, \emph{e.g.}, $L2$-distance; $[m + \cdot ]_{+}$ denotes the hinge function; $C$ denotes the number of semantic labels.

In this way, the Semantic Triplet Loss builds the relationship of various descriptors based on semantic information.
It supervises the model to discover semantically matching correspondences across images and further improves the performance for RGB-NIR cross-modality image registration.

\textbf{Overall loss function.}
The overall loss function of SGFormer is a weighted sum of three parts, including Semantic Triplet Loss ($\mathcal{L}_{st}$), Coarse-Matching Loss ($\mathcal{L}_{c}$), and Fine-Matching Loss ($\mathcal{L}_{f}$).
Details are as follows.
\begin{equation}\small
\mathcal{L} = \lambda_1 \mathcal{L}_{st} + \lambda_2 \mathcal{L}_{c} + \lambda_3 \mathcal{L}_{f}   
\end{equation}
To calculate $\mathcal{L}_{c}$, we follow LoFTR \cite{sun2021loftr} to create the ground truth matrix ($M_{gt}$), and use the dual-softmax score matrix ($M_{ds}$) for coarse-level feature matching. 
$L_{c}$ is then supervised by the cross entropy loss:
\begin{equation}\small
\mathcal{L}_{c} = - \frac{1}{|M_{gt}|} \sum_{(i,j) \in M_{gt}} log(M_{ds}(i, j))
\end{equation}
Similar to \cite{sun2021loftr}, $\mathcal{L}_{f}$ minimizes the distance between the fine-matching prediction and the ground truth coordinates by $\ell_2$ loss.

\section{Experiment}

\subsection{Experiment Setting}


\textbf{Implementation Details}. We exploited a pre-trained semantic segmentation model from \cite{zhou2017scene,zhou2018semantic} to generate semantic labels for each patch.
The general encoder was based on the ResNet-18 \cite{he2016deep} backbone, which was similar to \cite{sun2021loftr,chen2022aspanformer}.
As for the training data, considering the requirement of the large volume of training images for image registration and comparisons with other image registration methods, we learned our SGFormer on visible images and tested the performance on cross-modality image registration.
Specifically, our model was trained on the MegaDepth \cite{li2018megadepth} dataset with the batch size of 32 and the image size of 640 $\times$ 640.
We used 32 RTX 2080Ti to train the model, with Adam optimizer \cite{adam} and the learning rate of $1\times 10^{-3}$.
We set the number of total epochs to 200, each of which was comprised of 512 randomly selected mini-batches.
For hyperparameters, we set $T=100$, $C=150$, $\lambda_1 = 1\times 10^{-2}$, $\lambda_2 = 1$, and $\lambda_3 = 1$.

\subsection{Homography Estimation}

\begin{table}[t]
\caption{\textbf{Homography estimation on proposed RGB-NIR-IRegis cross-modality benchmark}, measured in AUC (higher is better).  SGFormer outperformed the state-of-the-art methods by a large margin, including both general and cross-modality image registration methods.}
\begin{center}
    \renewcommand{\arraystretch}{1}
    \setlength{\tabcolsep}{8pt}
    \resizebox{0.98\textwidth}{!}{  
    \begin{tabular}{l| c|ccc|ccc}
    \toprule
     \multirow{2}{*}{Category} & \multirow{2}{*}{Method} &  \multicolumn{3}{c|}{Without viewpoint variations} & \multicolumn{3}{c}{With viewpoint variations} \\ \cline{3-8}
     & & $@3$px & $@5$px & $@10$px & $@3$px & $@5$px & $@10$px \\
        \midrule
    \multirow{3}{*}{Traditional} & HOG \cite{dalal2005histograms} & 19.8 & 31.4 & 46.3 & 0 & 0 & 0 \\
        & ORB \cite{rublee2011orb}   & 22.7 & 33.7 & 48.0 & 0.5 & 0.6 & 0.7 \\
        & SIFT \cite{lowe1999object} & 42.3 & 54.3 & 66.5 & 0.6 & 0.7 & 0.7\\
    \midrule
      \multirow{4}{*}{General} & SuperPoint \cite{detone2018superpoint} & 65.4 & 78.4 & 88.7 & 11.6 & 21.6 & 38.8  \\
       & SuperGlue \cite{sarlin2020superglue} & 66.0 & 78.7 & 88.8 & 10.8 & 19.9 & 36.7 \\
       & LoFTR \cite{sun2021loftr}  & 79.2 & 87.4 & 93.7 & 13.9 & 23.5 & 39.3\\
      & ASpanFormer \cite{chen2022aspanformer} & 78.8 & 87.2 & 93.6 & 14.0 & 23.4 & 39.2 \\
    \midrule
    \multirow{3}{*}{Cross-modality} & DASC \cite{kim2016dasc} & 56.9 & 74.2 & 87.1 & 0.6 & 0.7 & 0.7 \\
    & LGHD \cite{lghd2015} & 43.9  & 57.1 & 70.7 & 6.5 & 12.5 & 22.6 \\
    & \textbf{SGFormer(Ours)}  & \textbf{82.2}($\uparrow$3.0)  & \textbf{89.3}($\uparrow$1.9) & \textbf{94.7}($\uparrow$1.0) & \textbf{15.8}($\uparrow$1.9) & \textbf{25.8}($\uparrow$2.3) & \textbf{41.1}($\uparrow$1.8) \\
    \bottomrule
    \end{tabular}
    }
\end{center}
\label{tab:RGB-NIR Matching}
\end{table}

\begin{figure*}[t]
    \centering
\includegraphics[width=0.8\textwidth]{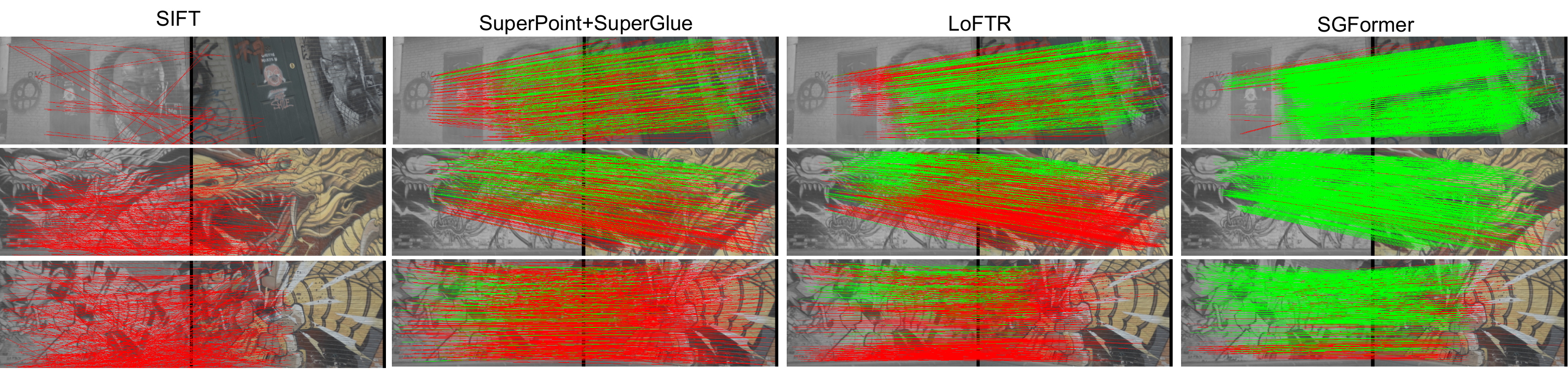}
\caption{\textbf{Qualitative comparisons on RGB-NIR image pairs with viewpoint variations in our RGB-NIR-IRegis benchmark.}
Results show that our proposed SGFormer was able to correctly match the correspondences in this challenging scenario, which outperformed previous methods.
}
\label{fig: vis comparsion result}
\end{figure*}

\textbf{Cross-modality Image Registration.} We first evaluated previous registration methods on our RGB-NIR-IRegis benchmark for cross-modality homography estimation.
Specifically, given the matching correspondences generated from our method, we adopted OpenCV to calculate the homography estimation with RANSAC.
Following \cite{sun2021loftr}, the number of output matches was set to be no more than 1K and all images were resized with the shorter side equal to 480.
We then reported the AUC of the corner error with the thresholds of 3, 5, and 10, which were widely exploited in previous studies \cite{sarlin2020superglue,sun2021loftr}.

\begin{table}[t]
\caption{\textbf{Homography estimation on HPatches (RGB-RGB) \cite{balntas2017hpatches}}, measured in AUC. Results demonstrate that high-level semantic information also provided beneficial guidance for the task of general image registration.}
\begin{center}
    \renewcommand{\arraystretch}{1}
    \setlength{\tabcolsep}{10pt}
    \resizebox{0.8\textwidth}{!}{  
    \begin{tabular}{l| cccc}
        \toprule
        Category &  Method  & $@3$px & $@5$px & $@10$px\\
       \midrule
       \multirow{7}{*}{General} &    SuperPoint \cite{detone2018superpoint} & 50.3 & 64.7 & 78.6 \\
        &R2D2+NN \cite{revaud2019r2d2} & 50.6 & 63.9 & 76.8 \\ 
        &SuperGlue \cite{sarlin2020superglue} &  53.9 & 68.3 & 81.7 \\
        &Patch2Pix \cite{zhou2021patch2pix} & 59.3 & 70.6 & 81.2 \\
        &LoFTR \cite{sun2021loftr}  &  65.9 & 75.6 & 84.6 \\
        &3DG-STFM \cite{mao20223dg}  &  64.7 & 73.1 & 81.0 \\
        &ASpanFormer \cite{chen2022aspanformer} & 65.4 & 75.8 & 85.2 \\
        \midrule
        \multirow{3}{*}{Cross-modality} & 
        DASC \cite{kim2016dasc} & 31.6 & 38.3 & 43.4 \\
        & LGHD \cite{lghd2015} & 27.6  &  37.4  & 50.3 \\
        & \textbf{SGFormer(Ours)} & \textbf{67.0}($\uparrow$1.1) & \textbf{76.6}($\uparrow$0.8)  & \textbf{85.6}($\uparrow$0.4) \\
        \bottomrule
    \end{tabular}
    }
\end{center}
\label{tab:hpatches}
\end{table}

\begin{table}[t]
\caption{
\textbf{Ablation study} on the efficacy of the proposed modules in SGFormer.
}
\begin{center}
\renewcommand{\arraystretch}{1}
\setlength{\tabcolsep}{10pt}
\resizebox{0.98\textwidth}{!}{  
\begin{tabular}{l |lll| lll}
\toprule
    \multirow{2}{*}{ Method }       &  \multicolumn{3}{|c|}{Without viewpoint variations}  & \multicolumn{3}{c}{With viewpoint variations}  \\ \cline{2-7}
     & $@3$px & $@5$px & $@10$px & $@3$px & $@5$px & $@10$px \\
   \midrule
   without Semantic Injection Module (SIM) & 79.2 & 87.4 & 93.7 & 13.9 & 23.5 & 39.3\\
   without Semantic Triplet Loss (STL) & 81.4 & 88.8 & 94.4 & 14.8 & 25.1 & 40.6 \\
   with SIM and STL & \textbf{82.2} & \textbf{89.3} & \textbf{94.7}  & \textbf{15.8} & \textbf{25.8}  & \textbf{41.1} \\
\bottomrule
\end{tabular}
}
\end{center}
\label{tab:Ablation Study}
\end{table}


Results are summarised in Table \ref{tab:RGB-NIR Matching}. 
To this end, it is noticeable that current registration methods performed poorly on our proposed RGB-NIR-IRegis benchmark, which underscores its significant challenges for the task of RGB-NIR cross-modality image registration.
In contrast, despite the simplicity of the design, our baseline method SGFormer successfully outperformed previous methods under all AUC thresholds by a large margin and achieved a new state-of-the-art performance for RGB-NIR cross-modality image registration, especially on unaligned image pairs, \emph{e.g.}, \textbf{0.7\% (SIFT)} \textit{\textbf{vs}} \textbf{39.3\% (LoFTR)}  \textit{\textbf{vs}} \textbf{22.6\% (LGHD)} 
 \textit{\textbf{vs}} \textbf{41.1\% (Ours)}.
More qualitative results are shown in Fig. \ref{fig: vis comparsion result}.
We attribute the performance gain to the design of Semantic Injection Module (SIM), which \textit{globally} integrates high-level semantic information into descriptors with Eq. \ref{EQ:SIM}.
In addition, Semantic Triplet Loss also contributes to the accurate estimation by forcing \textit{local} patches
to have semantically-consistent descriptors.


\textbf{General Image Registration.}
To comprehensively evaluate our proposed SGFormer, we also conduct the homography estimation experiment on HPatches \cite{balntas2017hpatches} for general image registration.
We followed the same evaluation protocol in our cross-modality homography estimation experiment.
As shown in Table \ref{tab:hpatches}, SGFormer achieved the best performance, indicating the value of high-level semantic information for general image registration.
\textbf{Ablation Study}. 
We further investigated the efficacy of several key components in our SGFormer. 
As shown in Table \ref{tab:Ablation Study}, when combining both the Semantic Injection Module and Semantic Triplet Loss, our method achieved the best performance on both aligned and unaligned RGB-NIR image pairs in our RGB-NIR-IRegis benchmark for homography estimation.
Such results demonstrate the effectiveness of the proposed modules in our SGFormer.

\section{Conclusion}

In this paper, we have focused on the area of RGB-NIR cross-modality image registration, which is crucial for many downstream vision tasks to fully utilize the complementary information in visible and infrared images. 
To first ensure a rigorous investigation, we have presented the RGB-NIR-IRegis benchmark, which, to the best of our knowledge, is the largest-scale RGB-NIR cross-modality image registration benchmark to date, featuring pixel-wise correct annotations and significant viewpoint variations for the first time. 
Based on the RGB-NIR-IRegis benchmark, we have fairly shown that the performance of previous methods is far from satisfactory, especially on RGB-NIR image pairs with viewpoint variations, which highlights the significant challenges posed by our benchmark. 
To analyze the causes of the poor performance, we have conducted thorough analyses that revealed the negative impact of inconsistent local features between RGB-NIR image pairs on the model performance.
These analyses have further motivated us to propose a baseline method, named the Semantic Guidance Transformer (SGFormer), which achieved a new state-of-the-art performance on the task of RGB-NIR cross-modality image registration despite the simplicity of its design.

Nevertheless, we also admit that like any benchmarks and methods, our research is not without any limitations.
The geometric differences in our benchmark may not be fully diverse for sufficient evaluations. 
And the introduction of semantic information may not be the best way to fully solve the issue of this field.
However, we believe that it should be \textit{necessary} for future studies to achieve strong performance on our benchmark to demonstrate their broad and robust capabilities.
And we expect to develop stronger models in future work, hoping to invite more attention from researchers to revisit this critical domain.
%
%
\bibliographystyle{splncs04}
\bibliography{main}

\begin{thebibliography}{10}
\providecommand{\url}[1]{\texttt{#1}}
\providecommand{\urlprefix}{URL }
\providecommand{\doi}[1]{https://doi.org/#1}

\bibitem{lghd2015}
Aguilera, C., Sappa, A.D., Toledo, R.: Lghd: A feature descriptor for matching across non-linear intensity variations. In: Image Processing (ICIP), 2015 IEEE International Conference on. p.~5. IEEE (Sep 2015)

\bibitem{arar2020unsupervised}
Arar, M., Ginger, Y., Danon, D., Bermano, A.H., Cohen-Or, D.: Unsupervised multi-modal image registration via geometry preserving image-to-image translation. In: Proceedings of the IEEE/CVF conference on computer vision and pattern recognition. pp. 13410--13419 (2020)

\bibitem{balntas2017hpatches}
Balntas, V., Lenc, K., Vedaldi, A., Mikolajczyk, K.: Hpatches: A benchmark and evaluation of handcrafted and learned local descriptors. In: Proceedings of the IEEE conference on computer vision and pattern recognition. pp. 5173--5182 (2017)

\bibitem{brown2011multi}
Brown, M., S{\"u}sstrunk, S.: Multi-spectral sift for scene category recognition. In: CVPR 2011. pp. 177--184. IEEE (2011)

\bibitem{chen2022aspanformer}
Chen, H., Luo, Z., Zhou, L., Tian, Y., Zhen, M., Fang, T., McKinnon, D., Tsin, Y., Quan, L.: Aspanformer: Detector-free image matching with adaptive span transformer. In: European Conference on Computer Vision. pp. 20--36. Springer (2022)

\bibitem{chen2020simswap}
Chen, R., Chen, X., Ni, B., Ge, Y.: Simswap: An efficient framework for high fidelity face swapping. In: Proceedings of the 28th ACM International Conference on Multimedia. pp. 2003--2011 (2020)

\bibitem{chen2022degradation}
Chen, X., Xiong, Z., Cheng, Z., Peng, J., Zhang, Y., Zha, Z.J.: Degradation-agnostic correspondence from resolution-asymmetric stereo. In: Proceedings of the IEEE/CVF Conference on Computer Vision and Pattern Recognition. pp. 12962--12971 (2022)

\bibitem{chen2022guide}
Chen, Y., Huang, D., Xu, S., Liu, J., Liu, Y.: Guide local feature matching by overlap estimation. arXiv preprint arXiv:2202.09050  (2022)

\bibitem{dalal2005histograms}
Dalal, N., Triggs, B.: Histograms of oriented gradients for human detection. In: 2005 IEEE computer society conference on computer vision and pattern recognition (CVPR'05). vol.~1, pp. 886--893. Ieee (2005)

\bibitem{detone2018superpoint}
DeTone, D., Malisiewicz, T., Rabinovich, A.: Superpoint: Self-supervised interest point detection and description. In: Proceedings of the IEEE conference on computer vision and pattern recognition workshops. pp. 224--236 (2018)

\bibitem{edstedt2022dkm}
Edstedt, J., Athanasiadis, I., Wadenb{\"a}ck, M., Felsberg, M.: Dkm: Dense kernelized feature matching for geometry estimation. arXiv preprint arXiv:2202.00667  (2022)

\bibitem{he2016deep}
He, K., Zhang, X., Ren, S., Sun, J.: Deep residual learning for image recognition. In: Proceedings of the IEEE conference on computer vision and pattern recognition. pp. 770--778 (2016)

\bibitem{heo2010robust}
Heo, Y.S., Lee, K.M., Lee, S.U.: Robust stereo matching using adaptive normalized cross-correlation. IEEE Transactions on pattern analysis and machine intelligence  \textbf{33}(4),  807--822 (2010)

\bibitem{hermans2017defense}
Hermans, A., Beyer, L., Leibe, B.: In defense of the triplet loss for person re-identification. arXiv preprint arXiv:1703.07737  (2017)

\bibitem{huang2017arbitrary}
Huang, X., Belongie, S.: Arbitrary style transfer in real-time with adaptive instance normalization. In: Proceedings of the IEEE international conference on computer vision. pp. 1501--1510 (2017)

\bibitem{irani1998robust}
Irani, M., Anandan, P.: Robust multi-sensor image alignment. In: Sixth International Conference on Computer Vision (IEEE Cat. No. 98CH36271). pp. 959--966. IEEE (1998)

\bibitem{jiang2021cotr}
Jiang, W., Trulls, E., Hosang, J., Tagliasacchi, A., Yi, K.M.: Cotr: Correspondence transformer for matching across images. In: Proceedings of the IEEE/CVF International Conference on Computer Vision. pp. 6207--6217 (2021)

\bibitem{jin2022darkvisionnet}
Jin, S., Yu, B., Jing, M., Zhou, Y., Liang, J., Ji, R.: Darkvisionnet: Low-light imaging via rgb-nir fusion with deep inconsistency prior. In: Proceedings of the AAAI Conference on Artificial Intelligence. pp. 1104--1112 (2022)

\bibitem{kim2016dasc}
Kim, S., Min, D., Ham, B., Do, M.N., Sohn, K.: Dasc: Robust dense descriptor for multi-modal and multi-spectral correspondence estimation. IEEE transactions on pattern analysis and machine intelligence  \textbf{39}(9),  1712--1729 (2016)

\bibitem{kim2015dasc}
Kim, S., Min, D., Ham, B., Ryu, S., Do, M.N., Sohn, K.: Dasc: Dense adaptive self-correlation descriptor for multi-modal and multi-spectral correspondence. In: Proceedings of the IEEE conference on computer vision and pattern recognition. pp. 2103--2112 (2015)

\bibitem{adam}
Kingma, D.P., Ba, J.: Adam: A method for stochastic optimization. In: ICLR (Poster) (2015)

\bibitem{faceshifter}
Li, L., Bao, J., Yang, H., Chen, D., Wen, F.: Advancing high fidelity identity swapping for forgery detection. In: Proceedings of the IEEE/CVF Conference on Computer Vision and Pattern Recognition. pp. 5074--5083 (2020)

\bibitem{li2018megadepth}
Li, Z., Snavely, N.: Megadepth: Learning single-view depth prediction from internet photos. In: Proceedings of the IEEE conference on computer vision and pattern recognition. pp. 2041--2050 (2018)

\bibitem{liang2019unsupervised}
Liang, M., Guo, X., Li, H., Wang, X., Song, Y.: Unsupervised cross-spectral stereo matching by learning to synthesize. In: Proceedings of the AAAI Conference on Artificial Intelligence. vol.~33, pp. 8706--8713 (2019)

\bibitem{liang2022deep}
Liang, X., Jung, C.: Deep cross spectral stereo matching using multi-spectral image fusion. IEEE Robotics and Automation Letters  \textbf{7}(2),  5373--5380 (2022)

\bibitem{objectdetection}
Liu, J., Fan, X., Huang, Z., Wu, G., Liu, R., Zhong, W., Luo, Z.: Target-aware dual adversarial learning and a multi-scenario multi-modality benchmark to fuse infrared and visible for object detection. In: Proceedings of the IEEE/CVF Conference on Computer Vision and Pattern Recognition. pp. 5802--5811 (2022)

\bibitem{lowe1999object}
Lowe, D.G.: Object recognition from local scale-invariant features. In: Proceedings of the seventh IEEE international conference on computer vision. vol.~2, pp. 1150--1157. Ieee (1999)

\bibitem{mao20223dg}
Mao, R., Bai, C., An, Y., Zhu, F., Lu, C.: 3dg-stfm: 3d geometric guided student-teacher feature matching. In: European Conference on Computer Vision. pp. 125--142. Springer (2022)

\bibitem{Martin2019DriveActAM}
Martin, M., Roitberg, A., Haurilet, M., Horne, M., Rei{\ss}, S., Voit, M., Stiefelhagen, R.: Drive\&act: A multi-modal dataset for fine-grained driver behavior recognition in autonomous vehicles. 2019 IEEE/CVF International Conference on Computer Vision (ICCV) pp. 2801--2810 (2019)

\bibitem{mishkin2018repeatability}
Mishkin, D., Radenovic, F., Matas, J.: Repeatability is not enough: Learning affine regions via discriminability. In: Proceedings of the European Conference on Computer Vision (ECCV). pp. 284--300 (2018)

\bibitem{surveillance}
Paramanandham, N., Rajendiran, K.: Infrared and visible image fusion using discrete cosine transform and swarm intelligence for surveillance applications. Infrared Physics \& Technology  \textbf{88},  13--22 (2018)

\bibitem{SPADE}
Park, T., Liu, M.Y., Wang, T.C., Zhu, J.Y.: Semantic image synthesis with spatially-adaptive normalization. 2019 IEEE/CVF Conference on Computer Vision and Pattern Recognition (CVPR) pp. 2332--2341 (2019)

\bibitem{revaud2019r2d2}
Revaud, J., Weinzaepfel, P., De~Souza, C., Pion, N., Csurka, G., Cabon, Y., Humenberger, M.: R2d2: repeatable and reliable detector and descriptor. arXiv preprint arXiv:1906.06195  (2019)

\bibitem{rublee2011orb}
Rublee, E., Rabaud, V., Konolige, K., Bradski, G.: Orb: An efficient alternative to sift or surf. In: 2011 International conference on computer vision. pp. 2564--2571. Ieee (2011)

\bibitem{Salamati2012SemanticIS}
Salamati, N., Larlus, D., Csurka, G., S{\"u}sstrunk, S.: Semantic image segmentation using visible and near-infrared channels. In: ECCV Workshops (2012)

\bibitem{sarlin2020superglue}
Sarlin, P.E., DeTone, D., Malisiewicz, T., Rabinovich, A.: Superglue: Learning feature matching with graph neural networks. In: Proceedings of the IEEE/CVF conference on computer vision and pattern recognition. pp. 4938--4947 (2020)

\bibitem{shen2014multi}
Shen, X., Xu, L., Zhang, Q., Jia, J.: Multi-modal and multi-spectral registration for natural images. In: European Conference on Computer Vision. pp. 309--324. Springer (2014)

\bibitem{sun2021loftr}
Sun, J., Shen, Z., Wang, Y., Bao, H., Zhou, X.: Loftr: Detector-free local feature matching with transformers. In: Proceedings of the IEEE/CVF conference on computer vision and pattern recognition. pp. 8922--8931 (2021)

\bibitem{autonomousdriving}
Takumi, K., Watanabe, K., Ha, Q., Tejero-De-Pablos, A., Ushiku, Y., Harada, T.: Multispectral object detection for autonomous vehicles. In: Proceedings of the on Thematic Workshops of ACM Multimedia 2017. pp. 35--43 (2017)

\bibitem{tang2015high}
Tang, H., Zhang, X., Zhuo, S., Chen, F., Kutulakos, K.N., Shen, L.: High resolution photography with an rgb-infrared camera. In: 2015 IEEE International Conference on Computational Photography (ICCP). pp. 1--10. IEEE (2015)

\bibitem{tosi2022rgb}
Tosi, F., Ramirez, P.Z., Poggi, M., Salti, S., Mattoccia, S., Di~Stefano, L.: Rgb-multispectral matching: Dataset, learning methodology, evaluation. In: Proceedings of the IEEE/CVF Conference on Computer Vision and Pattern Recognition. pp. 15958--15968 (2022)

\bibitem{walters2021there}
Walters, C., Mendez, O., Johnson, M., Bowden, R.: There and back again: Self-supervised multispectral correspondence estimation. In: 2021 IEEE International Conference on Robotics and Automation (ICRA). pp. 5147--5154. IEEE (2021)

\bibitem{wang2020learning}
Wang, Q., Zhou, X., Hariharan, B., Snavely, N.: Learning feature descriptors using camera pose supervision. In: European Conference on Computer Vision. pp. 757--774. Springer (2020)

\bibitem{xu2022rfnet}
Xu, H., Ma, J., Yuan, J., Le, Z., Liu, W.: Rfnet: Unsupervised network for mutually reinforcing multi-modal image registration and fusion. In: Proceedings of the IEEE/CVF conference on computer vision and pattern recognition. pp. 19679--19688 (2022)

\bibitem{facecontroller}
Xu, Z., Yu, X., Hong, Z., Zhu, Z., Han, J., Liu, J., Ding, E., Bai, X.: Facecontroller: Controllable attribute editing for face in the wild. In: AAAI Conference on Artificial Intelligence (2021)

\bibitem{Scalemap}
Yan, Q., Shen, X., Xu, L., Zhuo, S., Zhang, X., Shen, L., Jia, J.: Cross-field joint image restoration via scale map. In: Proceedings of the IEEE International Conference on Computer Vision. pp. 1537--1544 (2013)

\bibitem{reID1}
Yang, M., Huang, Z., Hu, P., Li, T., Lv, J., Peng, X.: Learning with twin noisy labels for visible-infrared person re-identification. In: Proceedings of the IEEE/CVF Conference on Computer Vision and Pattern Recognition. pp. 14308--14317 (2022)

\bibitem{Zhang2019FeatherNetsCN}
Zhang, P., Zou, F., Wu, Z., Dai, N., Mark, S., Fu, M., Zhao, J., Li, K.: Feathernets: Convolutional neural networks as light as feather for face anti-spoofing. 2019 IEEE/CVF Conference on Computer Vision and Pattern Recognition Workshops (CVPRW) pp. 1574--1583 (2019)

\bibitem{reID2}
Zhang, Q., Lai, C., Liu, J., Huang, N., Han, J.: Fmcnet: Feature-level modality compensation for visible-infrared person re-identification. In: Proceedings of the IEEE/CVF Conference on Computer Vision and Pattern Recognition. pp. 7349--7358 (2022)

\bibitem{Zhang2018ADA}
Zhang, S., Wang, X., Liu, A., Zhao, C., Wan, J., Escalera, S., Shi, H., Wang, Z., Li, S.: A dataset and benchmark for large-scale multi-modal face anti-spoofing. 2019 IEEE/CVF Conference on Computer Vision and Pattern Recognition (CVPR) pp. 919--928 (2018)

\bibitem{zhi2018deep}
Zhi, T., Pires, B.R., Hebert, M., Narasimhan, S.G.: Deep material-aware cross-spectral stereo matching. In: Proceedings of the IEEE conference on computer vision and pattern recognition. pp. 1916--1925 (2018)

\bibitem{Zhi2019MultispectralIF}
Zhi, T., Pires, B.R., Hebert, M., Narasimhan, S.G.: Multispectral imaging for fine-grained recognition of powders on complex backgrounds. 2019 IEEE/CVF Conference on Computer Vision and Pattern Recognition (CVPR) pp. 8691--8700 (2019)

\bibitem{zhou2017scene}
Zhou, B., Zhao, H., Puig, X., Fidler, S., Barriuso, A., Torralba, A.: Scene parsing through ade20k dataset. In: Proceedings of the IEEE Conference on Computer Vision and Pattern Recognition (2017)

\bibitem{zhou2018semantic}
Zhou, B., Zhao, H., Puig, X., Xiao, T., Fidler, S., Barriuso, A., Torralba, A.: Semantic understanding of scenes through the ade20k dataset. International Journal on Computer Vision  (2018)

\bibitem{zhou2021patch2pix}
Zhou, Q., Sattler, T., Leal-Taixe, L.: Patch2pix: Epipolar-guided pixel-level correspondences. In: Proceedings of the IEEE/CVF conference on computer vision and pattern recognition. pp. 4669--4678 (2021)

\bibitem{zhou2022promoting}
Zhou, S., Tan, W., Yan, B.: Promoting single-modal optical flow network for diverse cross-modal flow estimation. In: Proceedings of the AAAI Conference on Artificial Intelligence. vol.~36, pp. 3562--3570 (2022)

\end{thebibliography}
\end{document}